# FIRST LESSONS LEARNED OF AN ARTIFICIAL INTELLIGENCE ROBOTIC SYSTEM FOR AUTONOMOUS COARSE WASTE RECYCLING USING MULTISPECTRAL IMAGING-BASED METHODS

Timo Lange [1,2], Ajish Babu [4], Philipp Meyer [1], Matthis Keppner [1], Tim Tiedemann [1], Martin Wittmaier [3], Sebastian Wolff [3] and Thomas Vögele [4]

[1] Hamburg University of Applied Sciences, Berliner Tor 7, 20099 Hamburg, Germany
[2] School of Computing, Engineering and Physical Sciences, University of the West of Scotland, High St., Paisley PA1 2BE, UK
[3] Institute for Energy, Recycling and Environmental Protection at Bremen University of Applied Sciences., Neustadtswall 30, 28199 Bremen, Germany
[4] German Research Center for Artificial Intelligence, Robotics Innovation Center, Robert-Hooke Str 1, 28359 Bremen, Germany

ABSTRACT: Current disposal facilities for coarse-grained waste perform manual sorting of materials with heavy machinery. Large quantities of recyclable materials are lost to coarse waste, so more effective sorting processes must be developed to recover them. Two key aspects to automate the sorting process are object detection with material classification in mixed piles of waste, and autonomous control of hydraulic machinery. Because most objects in those accumulations of waste are damaged or destroyed, object detection alone is not feasible in the majority of cases. To address these challenges, we propose a classification of materials with multispectral images of ultraviolet (UV), visual (VIS), near infrared (NIR), and short-wave infrared (SWIR) spectrums. Solution for autonomous control of hydraulic heavy machines for sorting of bulky waste is being investigated using cost-effective cameras and artificial intelligence-based controllers.

*Keywords: Smart Recycling, Coarse Waste, Construction Waste, Multispectral Image Processing, Artificial Intelligence, Machine Learning, Robot Control, Deep Learning.*

## 1. INTRODUCTION

Material recycling is a vital part of the circular economy and global climate protection while also contributing to resource conservation. An essential prerequisite for material recycling is having sorted materials (paper, cardboard, plastics etc.). Sorting processes are mandatory and essential to further processing, prepare contained resources, and to offer them as secondary raw material on the market. This means that material cycles can be closed, which is becoming more and more important in the discussion about climate and environmental protection. For example, if waste, such as plastics, is recycled materially rather than energetically, climate damaging emissions can be reduced by 1,600 g to 2,000 g $CO_2$ eq./kg plastic waste (HDPE, LDPE, PET) (Rudolph et al., 2020). In contrast to landfilling, high-quality energy recovery can also contribute to climate protection as it results, for example in municipal waste, a reduction of emissions of approx. 422 g $CO_2$ eq./kg of waste (Wittmaier et al., 2009).

In any case, efficient sorting processes are a prerequisite.

Nowadays, small-sized waste can be sorted in conveyor belt systems where first artificial intelligence (AI) robotic systems are finding their way into such facilities, and the technology is continuously being developed (Zhang et al., 2019), which makes the sorting process more efficient. A wide variety of sensors are used to separate the different materials from each other (NIR, HIS, ISS, 3D etc.). However, the sorting process of small-sized waste nearly always takes place with belt support, with defined small distances between sensors and isolated waste material under almost constant external conditions.

In contrast, coarse-grained waste (e.g., construction and demolition waste, CDW) is still sorted by manually operated cranes, excavators, and other heavy machinery. As a result, large quantities of principally recyclable materials are lost to coarse waste (coarse waste in Germany: approx. 3 million Mg of bulky waste, 220 million Mg of construction and demolition waste etc.) (Destatis, 2022). Therefore, more effective processes for sorting coarse waste by type must be developed to recover recyclable materials from mixed waste.

Two key functions must be realized to enable an autonomous excavator for sorting of coarse waste: (1) Classification of objects and materials in the waste conglomerate. (2) Control of the manipulator and the gripper. Previous work describes the overall scope of our concept (Tiedemann et al., 2021) which is conducted in the research project SmartRecycling-UP.

Conditions of a sensor-based detection for automatic sorting of coarse-grained waste impose different challenges than sorting of small-sized waste via conveyor belt. On the one hand, the sensors cannot work in quasi-enclosed detection areas close by the waste objects and therefore must cope with influencing factors such as dust or changing lighting conditions, especially when detecting from the distance of, e.g., 10 m to 15 m. Also, the waste is not isolated before sorting, but is available as a heap. This leads to object overlaps, which makes it difficult to detect material and objects. Such data for material classification and object detection in heaps of trash are not yet been available but is of fundamental importance for the automation of these processes.

Regarding the control of the manipulator and the gripper, the technical environment and the business conditions often demand a solution which uses existing equipment. This could be stationary heavy machinery as hydraulically driven excavators. Controlling hydraulic machines can often face difficulties, for example, a varying, temperature-depending control stiffness. When autonomously controlling the excavator, the joint controller needs to compensate these variations as the human crane driver does it today.

In this work, we present the first lessons learned in the process to enable autonomous coarse waste sorting. To address these challenges, we propose machine learning (ML) based classification of materials with multispectral images of ultraviolet (UV), visual (VIS), near infrared (NIR), and short-wave infrared (SWIR) spectrums. Furthermore, we present our approach to automate retrofitted stationary and mobile hydraulic excavators with an artificial intelligence (AI) based control.

The reminder of this work organized as follows: Section 2 gives an overview of the overall system scope and presents its components and their tasks. Section 3 presents relevant related work. In Section 4, we discuss first results, findings and lessons learned while deploying the sensor related components. Afterwards, Section 5 elaborates on the corresponding results, findings and lessons learned in the progress of implementing an automated hydraulically driven excavator. Section 6 concludes the work with an outlook on future work.

## 2. OVERALL SYSTEM SCOPE

The concept presented in this paper is supposed to be integrated into an overall approach for an autonomous excavator. The approach proposes software modules realizing the control of the manipulator and the gripper as well as the detection and classification of objects and materials in the waste conglomerate.

Figure 1 shows the modules in the concept and their interconnections. The *SmartObjectClassifier* (see Section 4) uses multispectral imaging paired with AI to detect and classify materials in the heap of waste. It approaches the first sensory task which is the recognition of individual objects in the waste conglomerate. This task is planned as a two-step process where, first, known objects are identified in images taken by cameras installed in the facility. Second, for unknown objects, a material classification is carried out.

For the detected waste object classes, the *SmartObjectTracker* locates their pose. The resulting waste object list and poses are forwarded to the *SmartProcessController* which in turn takes input from the operator (e.g., sorting strategies, prioritization), and current pose of the robot, to compute the trajectory to move an object from the pit to the sorting location.

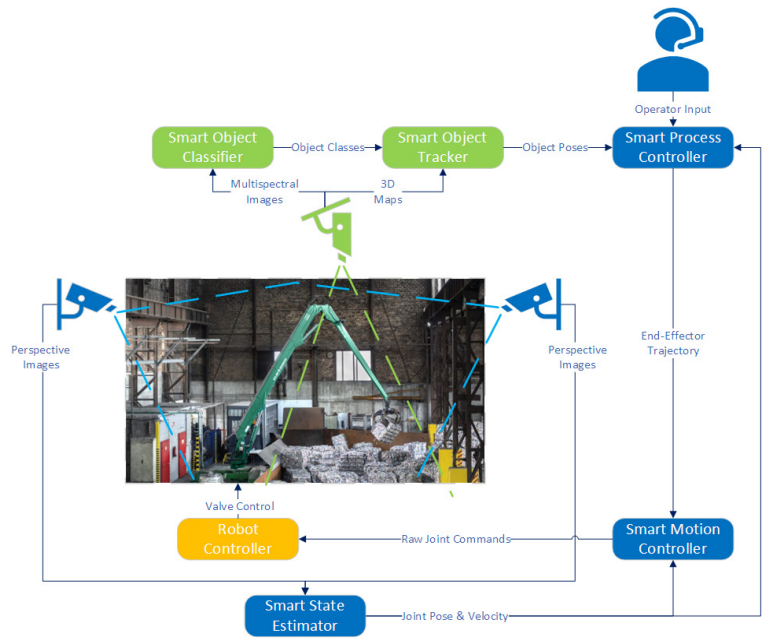

Figure 1. Concept developed in SmartRecycling-UP project for control of heavy hydraulic machines.

This trajectory is given to the *SmartMotionController* which executes the trajectory by using the estimated joint states from *SmartStateEstimator* and outputs the control output to the *RobotController*. The *RobotController* in turn controls the values based on the commands. The key components *SmartMotionController* (see Section 5.1) and *SmartStateEstimator* (see Section 5.2). Unlike the classical approach of explicitly modelling the dynamics of each joint, the proposed solution uses deep neural networks to learn the model, including the non-linearities. This learning requires the motion data of the joint during manual operation or automated data generation. The control concept also uses RGB cameras in place of the joint sensors for state estimation. The continuous sequence of images are processed by deep learning-based vision algorithms to estimate the position and velocity of the joints. These two innovations provide a solution that is generic, easy to set up and cost effective.

## 3. RELATED WORK

Techniques and applications of ML and deep learning (DL) are major topics of research in current and past research (Pouyanfar et al., 2018). An area in which DL algorithms achieve particularly promising results is the detection and classification of objects (Jiao et al., 2019). Due to the nature of disposal, most objects are heavily damaged when arriving in facilities that process coarse-grained waste. Object detection alone is thus not feasible in many cases.

A further possibility to make properties of objects visible for a computer is multispectral imaging (MSI) and hyperspectral imaging (HSI). Complex distinguishing features are usable when combining those images with DL (Huang et al.,2022; Jia et al., 2012).

ML and DL based technologies enable advanced systems in a multitude of operational areas like construction industry (Anding et al., 2013), food inspection (Qin et al., 2013) and recycling of waste (Kim et al., 2019). Procedures used include e.g. object detection, HSI, spectral imaging, depth images, SVMs, and DL. We evaluate the feasibility of combining MSI and ML in the application field of material classification in mixed coarse-grained waste.

Wang et al. (2020) present a mobile bulky waste classification. Their system is based on object detection in RGB images with a convolutional neural network (CNN). Most objects in recycling facilities are thrown together in large piles, are heavily damaged, and deformed which degrades the performance of object detection. Besides object recognition, other methods are needed to detect and distinguish contained materials.

Sakr et al. (2016) also use RGB images to classify waste into three types (plastic, paper, and metal). They compare the performance of SVM and CNN based classifiers. Their results show better detection results with SVM classification. We evaluate the feasibility of SVMs and DL based classification in seven types of materials (plastic, paper/cardboard, wood, metal, textiles, foam, and mineral/ stone). They are selected to cover a superset of all types of materials expected in the coarse-grained waste recycling facilities.

Kuritcyn et al. (2015) applied spectral analysis and ML to construction and demolition waste. They share the insight that RGB images need to be accompanied by further spectral information to get reasonable classification performance.

Using MS information to conclude compositions of observed objects is widely used in remote sensing applications (Toth & Jóźków, 2016; Zhou et al., 2018). Similar materials have to be distinguished at a much smaller distance in coarse waste recycling facilities.

Zhi et al. (2019) use MSI to classify powders. They reach 60%-70% accuracy in detecting the type of 100 different powders. A reasonable performance that we also observe even with limited training data and straightforward ML and DL procedures.

MS images of different cameras need to be registered to match image sections to get an accurate spectral resolution. Tosi et al. (2022) use a stereo-matching procedure to train the disparity of two RGB cameras. Their proposed deep network architecture is then also able to match RGB and MS camera images. Besides ML based matching procedures, one widely used approach for image matching is scale-invariant feature transform (SIFT) (Lowe, 1999). The SIFT algorithm is a feature-based image matching algorithm that can match images with different resolutions and distortions. The intensity of MS images is strongly fluctuating nonlinear and SIFT registration performance is degrading (Saleem & Sablatnig, 2014; Toth et al., 2010). There is past and recent work focusing on improving on SIFT (Saleem & Sablatnig, 2014; Toth et al., 2010) and its successor SURF (Jhan & Rau, 2021). Toth et al. (2010) note that SIFT is suitable for in-domain matching. We use SIFT to match the different cameras in their visible spectrum domain and then use this to also match the MS images outside of the shared visible domain.

Automated control of heavy hydraulic machines has been extensively addressed in the literature. The solution ranges from classical methods to using deep learning. The classical solution consists of retrofitting joints with sensors and explicit modeling of joints, parallel kinematic and non-linearities in the control loop, after which a Proportional-Integral-Derivative (PID) controller is used for velocity or position control of each joint. One such solution is described in (Babu et al., 2022). These methods have been extended with impedance control (hybrid position/force) of excavators such as (Ha et al., 2000). Recently usage of deep learning has been explored in control and excavation tasks (Egli et al., 2022; Egli & Hutter, 2022) as well. This work proposes a more extensive use of deep learning to attain a generic and cost effective solution for control of excavators and cranes, with focus on recycling of bulky waste.

## 4. CLASSIFICATION OF OBJECTS AND MATERIALS

For autonomous sorting of coarse waste, the classification of objects and material in the waste heap is a prerequisite with the strong demand to also identify interfering or hazardous materials. For object classification, detection and material classification different ML methods can be utilized. Current state of the art methods for object detection and classification in images are all based on deep learning (DL). While these methods archive remarkable results, they need huge amounts of training data. Although this is not a problem for common scenes and standard RGB images through techniques like transfer learning, it becomes an obstacle when little domain specific data is available and uncommon sensor modalities are used. When it comes to publicly available datasets of waste, these are solely comprising images of municipal waste captured with RGB cameras. So, there is a need to collect data of construction and demolition waste (CDW) and solid coarse grained industrial waste. Additionally, multiple spectrums must be recorded to investigate the usage of multispectral imaging-based methods as the most datasets available including multispectral or hyperspectral data are satellite images or recorded from unmanned aerial vehicle (UAV).

### 4.1 Data Acquisition/Collection

We collect at three kinds of multispectral (MS) datasets with slightly different purposes to refine the data collection process iteratively and incorporate intermediate findings. The first dataset is collected with a mobile sensor node for evaluation of effectiveness of different spectral bands for material classification and for initial training of straight forward ML models. Here, manual filter changes are performed, and automatic filter changes are tested. The data is collected in the lab and real environments. The second data collection is of larger size. Again, a mobile sensor node with automated spectral filter rotation captures trash in real environments. Depending on the first dataset evaluations, the second collection may contain a reduced selection of spectral bands. This dataset will be used to train sophisticated ML models, which will be enabled through the greater volume of data samples. The third dataset collection operates a sensor node directly attached to the ceiling above a waste bunker. This will also be the setup under which a future demonstrator is planned to operate.

In the data acquisition process, several steps are necessary to make the data suitable for usage with ML models. First the sensor setup and challenges encountered are described. Second, necessary preprocessing steps are presented. Finally, the data labeling process is discussed.

*4.1.1 Sensor Setup*

The sensor setup for multispectral imaging (MSI) is based on cameras in the UV, VIS/NIR and SWIR spectrums that cover an overall spectral range of 190nm to 1700nm. The filters are selected to partition parts of the full spectral range of all cameras. Table 1 shows the utilized filters for the corresponding cameras with the exposure times used in lab setting. Because the unfiltered VIS/NIR image is taken in RGB a resulting MS image contains 15 dimensions. The overlap in the visual spectrum of all cameras, with no filter attached, makes the calibration between sensors easier.

In the target environment, the sensors will be mounted to the ceiling above the heap of waste. Therefore, the cameras will capture images in a distance of approx. 3m to 10m depending on the height and location of the waste heap. To capture high resolution images with cost efficient sensors, 2D cameras are chosen with different bandpass filters. For the collection of different spectrums, the filters are either manually changed or by a filter wheel attached in front of the lens. Figure 2 shows the sensor node, which is used to initially record data for the first dataset collections.

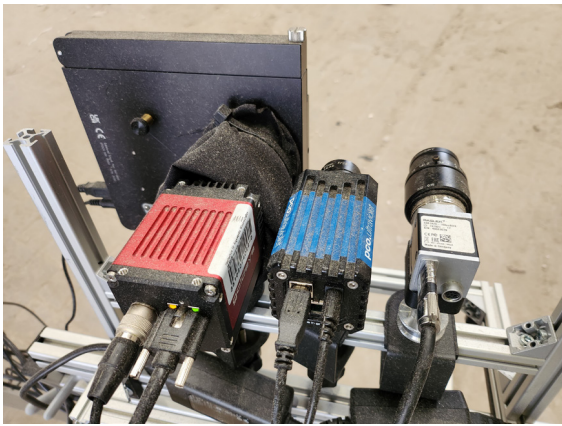

Figure 2. Sensor node with SWIR, UV and VIS cameras. A motorized filter wheel is mounted in front of the SWIR-camera lens for automatic optical filter changes.

For a less complex and expensive final hardware setup and less computational costs, one goal is to find decisive spectrums to reduce the number of required spectral bands as much as possible. First investigations indicate that UV spectrums where less significant for material classification. Without the UV spectrum the sensor node would be reduced to two cameras, but further evaluations have to be conducted to disqualify the UV spectrum.

A challenge for this sensor setup is the control of exposure time due to the largely uncontrollable environmental conditions like dust and sunlight. Since the image is constituted of the recordings from three different sensors and the images from one sensor are captured time-shifted, the sensor internal exposure time control cannot be used. Currently the exposure time is adapted manually to the present conditions. For complete automatic data acquisition, the control has to be automated. Current work evaluates approaches that use a calibration object in the Field of View (FoV), high dynamic range (HDR) images from exposure bracketing, an additional brightness sensor or adaption based on the mean image brightness.

In the final installation of the sensors, they must be placed in the same enclosure to minimize the distance to each other and to reduce the effort and error in image registration. Here, the wavelength to be transmitted must be considered and a protective pane with a wide light spectrum, like fused silica, has to be chosen.

Table 1. Overview of used optical filter configurations.

| Camera | Spectrum[nm] | Exposure time[s] | Index |
|---|---|---|---|
| UV | 190 - 1100 | 0.3 | 0 |
| | 290 - 365 | 5 | 1 |
| | 375 - 425<br>745 - 970 | 1 | 2 |
| VIS/NIR | 400 - 1000 | 0.01 | 3 |
| | 730 - 755 | 0.05 | 4 |
| | 830 - 865 | 0.1 | 5 |
| | 845 - 930 | 0.1 | 6 |
| | 928 - 955 | 0.4 | 7 |
| SWIR | 400 - 1700 | 0.04 | 8 |
| | 930 - 1030 | 0.4 | 9 |
| | 1290 - 1310 | 1 | 10 |
| | 1440 - 1460 | 1.5 | 11 |
| | 1485 - 1645 | 0.4 | 12 |

### 4.1.2 Preprocessing for Multispectral Images of Waste

The images of the cameras are subject to their respective image format and resolution. To remove objects that are not part of the waste and to further overlap the field of view of the cameras, the images are cropped to the area of interest. After that, the images are scaled to a common resolution for later merging. We use the robust and widely used SIFT algorithm (Lowe, 1999) to match the images of the cameras. FLANN-Matcher is used to find the best matches between the key points of the images. RANSAC is used to remove outliers from the matches. All three cameras cover the VIS spectrum (approx.400nm – 800nm). To bypass nonlinear intensity-based changes of MS images (Saleem & Sablatnig, 2014) we use VIS images of the cameras to search SIFT features and match their images. Figure 3 shows found features and matches in images from UV and VIS camera. To further improve the matching, we normalize each single image. This normalization is not included in the dataset. The dataset contains the original values. The matching itself is done by matching the VIS/NIR and SWIR cameras with the UV camera which look at the smallest area on the scene. The resulting images now contain 15 dimensions of MS data per pixel.

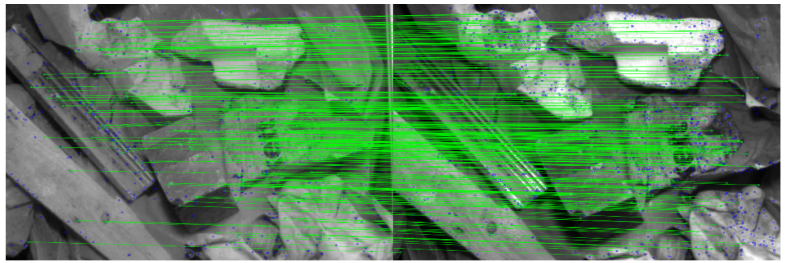

Figure 3. Camera SIFT matching in the visual spectrum without filter.

One problem with the calculation of the transformation per camera is that such slight changes in the position of the cameras within an exposure series, for example due to manual filter changes, are not calculated. This means that the overlay is no longer pixel precise. However, it is also not possible to calculate the transformation per image, because matching is not possible due to the different intensity values (Saleem & Sablatnig, 2014). To solve this problem, filter wheels will be used, which change the filters automatically. First tests show significantly less movement within a series.

*4.1.3 Data Labeling*

Most current object detection, classification and segmentation algorithms are based on supervised ML methods. Therefore, the recorded data must be annotated with class labels. Currently we distinguish seven classes (plastic, paper/cardboard, wood, metal, textiles, foam, and mineral/stone). For object detection/classification on the RGB channel, for human recognizable objects, we follow the standard approach with bounding boxes and class per image labels. For the per pixel material classification rectangles with a small margin to the edges in image sections with certain materials are drawn to account for inaccuracies in image registration and to avoid the time-consuming task of manual pixel-perfect segmentation labels. For later experiments with subcategories of plastics, a spectrometer was used during data collection to accurately determine the plastic type. Planned are precise pixelwise segmentation masks. Which poses a challenge due to the very high amount of manual work normally required for such a task. This requires a sophisticated pipeline to support the labelers. Preliminary tests with the Segment Anything Model (SAM) (Kirillov et al., 2023) showed promising results in automatically generating segmentation masks for deformed objects in the waste heap. SAM is a model which can generate masks for the whole image but without labels. Therefore, SAM can generate the masks, which is the most time-consuming part, and the manual work can be limited to correction of the masks and assignment of labels to the masks. Further challenges arise due to the data collection in real recycling facilities and no limitation to a fully controlled environment. It is not feasible to manually pick all objects from the waste heap for exact material classification and it is not possible for a human to distinguish all materials or objects solely from images after data collection. Therefore, not all objects and materials in the waste heap can be labeled.

## 4.2 Object Classification

Object detection and classification in images is a widely used technology in different domains. Heaps of coarse waste impose particular challenges on common detection and classification technologies.

The majority of objects are heavily deformed, dirty, shattered, and in unusual poses. Furthermore, they are all piled on top of each other with a large number of unidentifiable debris in between.

Figure 4 shows an example of a heap of waste that contain objects that are detectable for the human eye. In this case that objects are mattresses. All of them are partially covered, deformed, dirty and stuck between debris.

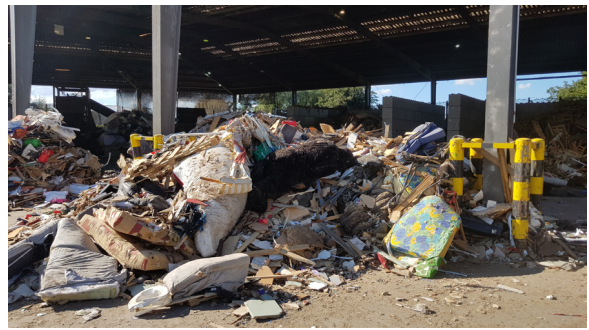

Figure 4. Heap of waste with mattresses in different pose, condition, and form.

First experiments indicated that up-to-date object detection methods and tools like YOLOv8[1], that are able locate and classify objects simultaneously, reach their limits in this particular conditions. The classification of individual objects on the other hand are feasible. Further solutions are needed to enable detection and classification.

Again, approaches that segment the image in the first step (e.g., SAM) show promising results. The prior segmentation filters unidentifiable debris, cut out the objects for individual classification, and also contains position of objects in the image.

## 4.3 Material Classification

Our initial approach for material classification is a per pixel classification where each pixel is individually classified by a multilayer perceptron (MLP) or support vector machine (SVM). The input consists of the multispectral channels of the individual pixels. Experiments with SVMs showed similar classification performance to the MLP but increased amounts of training samples led to excessive SVM training periods which makes them unusable for larger datasets.

---

[1] https://github.com/ultralytics/ultralytics

Figure 5 shows an example image of a waste scene with the per pixel classified image on the right and the corresponding RGB image on the left. Each color represents the predicted material class of the corresponding pixel. The scene is part of a small dataset with 27 scenes recorded in the lab under controlled conditions. This dataset was used to train an MLP. This example shows that the classification performance for paper/cardboard and foam (bottom left of the images) is low and shadow cast in gaps and on objects in the neighborhood leads to incorrect classifications. Additionally, dirt and adhesive residues on the objects pose a difficulty. In accordance with the resulting f1-score of 0.74 of our experiments the example also indicates that the approach is feasible.

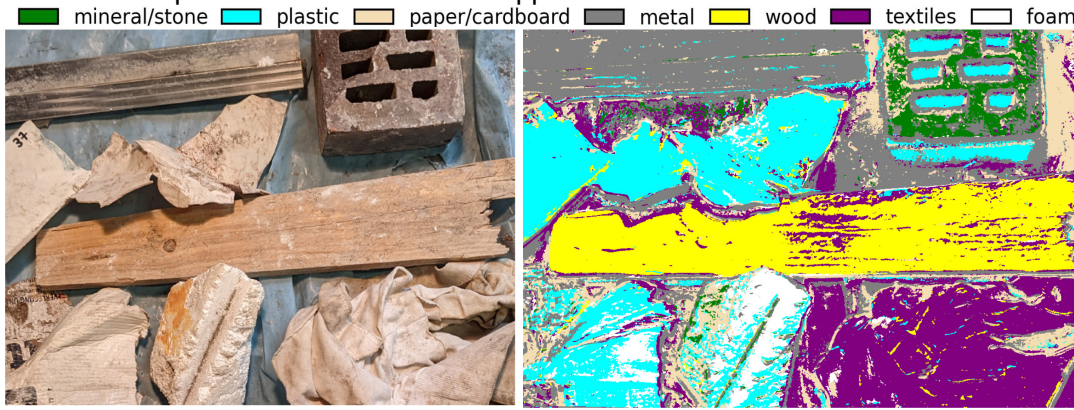

Figure 5. **Left:** RGB picture of a captured scene in lab setting. **Right:** Classified scene.

Figure 6 shows an example of classification for a scene in a recycling facility under uncontrolled conditions, which showcases the difficulty in classification under real conditions.
When data has been collected on a larger scale, advanced models like CNNs for semantic and instance segmentation, which require a greater amount of training data, will be evaluated.

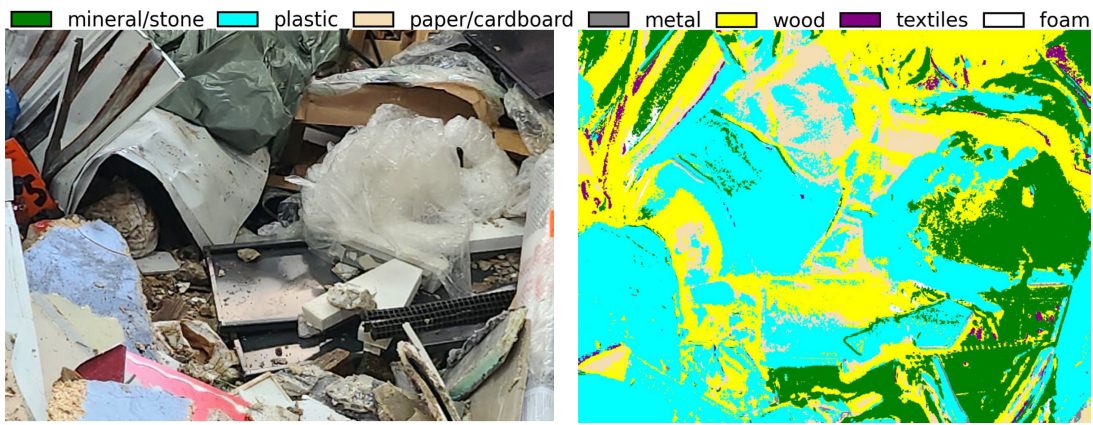

Figure 6. **Left:** RGB picture of a captured scene in recycling facility. **Right:** Classified scene.

### 4.4 Findings

Currently no publicly available multi or hyperspectral dataset of waste exists, which can be used to train ML methods for object detection, classification or segmentation. Additionally, there are no datasets of CDW available in the RGB spectrum. Therefore, to investigate the feasibility of multispectral and ML based systems not relying on conveyor belts, especially dealing with CDW, a dataset must be built up.

The annotation of the acquired data affords huge manual labor. Instance based segmentation without label through SAM is a promising possibility to support the labelers and reduce the manual labor substantially.

The challenge of exposure time adaption due to changing environmental conditions like dust and changing sunlight comes to the foreground in real environments. Possible solutions identified so far are a calibration object in the FoV, HDR images, an additional brightness sensor, or adaption based on the mean image brightness.

Another challenge is the image registration of MS images and from different sensors. During data collection with a mobile sensor node, the sensors may be subject to small changes in alignment or vibrations, which makes an initial calibration not possible. Currently the image registration leverages the fact that the sensors have an overlap in the VIS spectrum which bypass the problem of nonlinear intensity-based changes of MS images. Nevertheless, the registration is not yet pixel perfect and it presents an open problem which needs further investigation.

Current object detection methods have shown to not be able to sufficiently detect deformed, heavily covered, or soiled objects in the waste heap. An alternative solution would be to use general segmentation models like SAM, which cannot label objects but can be used to separate undefined instances of different objects, which can then be cut out and be classified with a separate model.

A relatively simple model like an MLP is already able to make use of different broad spectral bands for material classification and shows the feasibility of using a combination of multiple area cameras with a relatively high resolution and broad bands to directly classify materials in the waste heap compared to commonly used hyperspectral line scanners with a narrow spectral range and narrow spectral bands on a conveyer belt based system. A next step would be to extend the model to incorporate surrounding pixels to get more context for the classification. Finally, when a sufficiently large dataset is collected, more sophisticated models like CNNs or transformer-based segmentation models can be adapted to multiple spectrums and evaluated. An alternative solution for the usage of advanced models with a very small dataset is the usage of transfer learning from models trained on RGB to multispectral which will be investigated.

## 5. CONTROL OF HYDRAULIC HEAVY MACHINERY

Hydraulically actuated excavators and cranes are extensively used for handling waste, mainly due to their power and ruggedness. Automating such a machine requires accurate control of each joint which is controlled by hydraulic valves, and in some cases, multiple hydraulic valves (pilot control valve and main valve) for a single joint. In most cases, these valves are highly non-linear with varying dead-zones, non-linear oil flow, hysteresis, etc. Traditionally, for accurate control of the joints, these non-linearities should be modeled and the model parameters estimated.

Apart from this, the joints in most of these heavy machines have complex parallel kinematic mechanisms which determine the relationship between motion of the actuator and the joint. This also needs to be modeled and solved for forward and inverse kinematics for each type of mechanism.

An additional requirement is the joint position and velocity sensors for each joint to measure the state of the joint at every sampling time. This requires retrofitting of excavators and cranes with high-resolution sensors, which is expensive and time consuming.

Artificial intelligence, specifically deep learning-based solutions, can be used to address the above challenges and develop a solution which can be transferred to different hydraulic machines. It reduces the effort needed in retrofitting, modeling, and controller development. The key control components (c.f. Figure 1) are detailed in the following sections.

### 5.1 Smart Motion Controller (SMC)

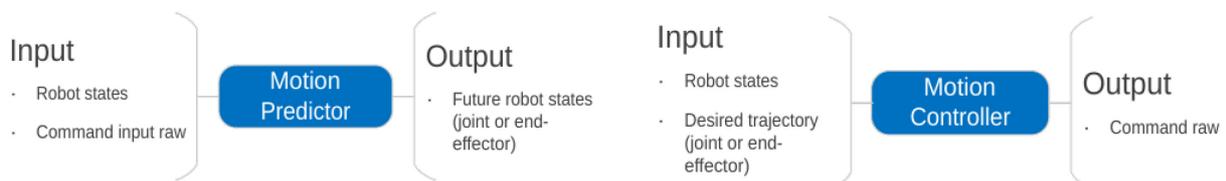

Figure 7. The two modules of SmartMotionController with their respective inputs and outputs.

The *SmartMotionController* is responsible for making the robot move in a particular trajectory by controlling the velocity of the joints. This requires two sub modules: *MotionPredictor* and *MotionController*.

The *MotionPredictor* tries to predict the motion of the robot based on the given raw commands and current robot state. The raw commands are the commands that are sent to the valves. The robot state consists of joint positions, current joint velocities, engine speed, hydraulic oil temperature, etc. The output of the *Motion Predictor* is the next robot state, as shown in figure 7 (left). This acts as a sort of simulation of the motion of the joints. This module is represented by a deep neural network and trained on data collected during the manual operation or by applying predefined motion profile.

The *MotionController* uses the learned *MotionPredictor* to simulate the dynamics of the robot and thereby learns a controller that makes the robot follow a trajectory accurately. Deep reinforcement learning based algorithms provide a powerful tool for learning the controller. The *MotionController* takes the current robot state and the desired trajectory as inputs and outputs the raw commands for the valves, as shown in figure 7 (right). The controller is also a deep neural network. The concept of *SmartMotionController* is similar to related work (Egli & Hutter, 2022).

## 5.2 Smart State Estimator (SSE)

The SmartStateEstimator estimates the state of the joints (position and velocity) based on the sensor input. A variety of sensors are applicable here including joint encoders, RGB cameras, depth cameras, lidars, motion tracking cameras, etc. Joint encoders are the most widely used and reliable solution, yet it is difficult to integrate or retrofit into an existing system. Depth cameras are expensive and the quality of depth image is poor at larger ranges. Lidars, in general, have lower resolution and are expensive. Similarly, motion tracking systems are very expensive and not usable for rugged applications.

As per our current assessment, the most efficient solution would be to use traditional RGB cameras and make use of recent advancements in deep learning for perception applications. Different solutions already exist in literature. Single object 6D pose estimation, as described in (Tekin et al., 2018), could be used to estimate the pose of each robot link and then compute the joint angle. Another possibility is to use networks trained using synthetic images to estimate joint angles and end-effector pose as discussed in (Zuo et al., 2019). This method has also been used to control a toy robot. Another method using synthetic images to estimate the camera-to-robot pose is developed in (Lee et al., 2019). These existing methods are currently being evaluated for performance and robustness. Apart from these existing methods, new methods also are being developed using implicit representations (Niemeyer et al., 2020) and skeleton tracking (Mehta et al., 2017).

## 5.3 Motion Predictor: Single Joint

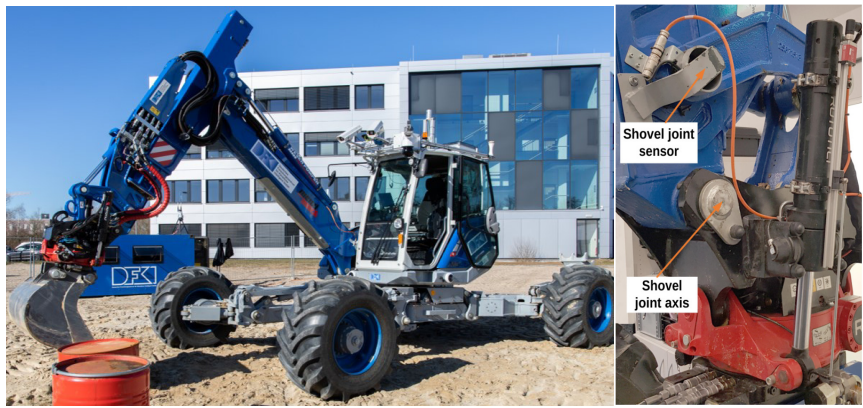

As a starting point, the *MotionPredictor* for a single joint was first learned and evaluated. The ARTER (Autonomous Rough Terrain Excavator Robot) platform that was developed by DFKI as part of the ROBDEKON project (Petereit et al., 2019) is used as the test platform. This platform, described in (Babu et al., 2022), is shown in figure 8 (left). The joint, *Shovel*, was used due to ease of operation and reduced chance of self or external collision. This joint is kinematically complex, and the joint encoder is attached to an axis which is part of the parallel kinematic chain, and not directly on the joint output axis. Figure 8 (right) shows the joint axis and the sensor attachment location for *Shovel* joint.

Figure 8. ARTER robot (left) and the robot's Shovel joint (right).

The objective of the *MotionPredictor* is to predict the motion of the joint based on the current state and the current command being sent to the valve. The current state consists of the sensor position, engine speed, oil temperature, and raw command for shovel joint. The predicted value is the velocity at the next time step. The data for learning was collected by applying a Chirp signal which is a sinusoidal signal with varying amplitude and frequency. The validation data was collected using raw commands sent from a joystick, which is manually operated.

The *MotionPrediction* network consists of a Long Short-Term Memory (LSTM) (Hochreiter & Schmidhuber, 1997) artificial neural network, which is a type of recurrent neural network (Yu et al., 2019) with an inbuilt mechanism for remembering old states. Figure 9 shows the prediction of the joint velocity for the training data and the figure 10 shows the prediction for the validation data. In both figures, the top plots are the actual and predicted joint velocity plotted against time. The bottom plots show the error between the actual and predicted velocities. The magnitude of the maximum errors in training and validation data are within 0.03 rad/s and 0.6 rad/s respectively.

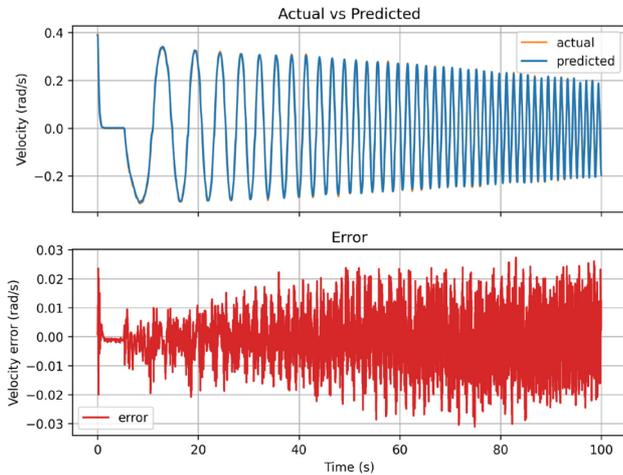

Figure 9. Performance of learned network for prediction of motion with training data.

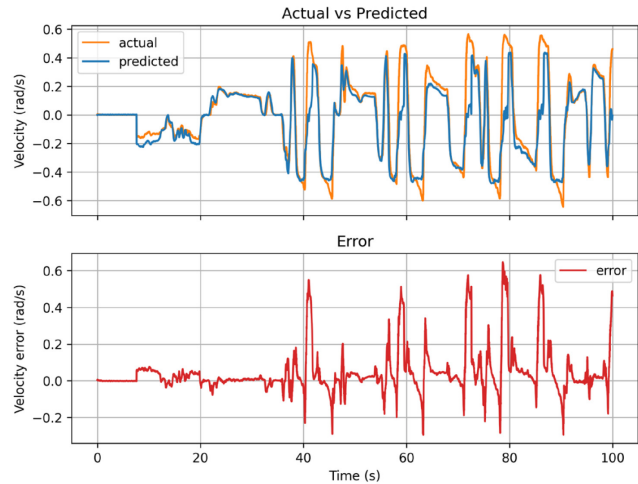

Figure 10. Performance of learned network for prediction of motion with test data.

### 5.4 Findings

In general, the network can model the dynamics of a joint. As can be seen from the plots (c.f. Figure 9 and Figure 10), the training data fits well. The learned model is able to represent the test data, but with higher inaccuracies. This is due to the limited amount of quality training data used for this learning. As can be seen from the training data, the velocity amplitudes are lesser compared to the test data. This results in fitting inaccuracies at high velocities. Another limitation with the training data is that the accelerations are far lesser compared to the test data which has higher accelerations. Hence the training data is unable to capture the full range of dynamics of the system. The above limitations can be remedied by taking more training data at higher velocities.

Since the training data is fitting well, it can be assumed that the network is able to approximate the nonlinearities of the system well. Since the input data contains only the sensor position, and not joint position or velocities, the network is able to incorporate the parallel kinematics and hydraulic non-linearities adequately.

To collect more data, with more dynamic range, an automated method of data collection needs to be developed. This should be extended for all joints simultaneously with self as well as external collision in mind.

### 6. CONCLUSIONS AND OUTLOOK

Our concept of an AI robotic system for autonomous coarse waste recycling is based on two key components: Object detection with material classification based on multispectral images, and autonomous control of hydraulic heavy machines.

For coarse grain waste sorting, a new solution is needed to detect objects and materials in a heap. In contrast to conveyer belt based systems, our approach uses a setup of three high resolution area scan cameras covering UV, VIS, NIR, and SWIR spectrums, whose images are registered into an MS image with 15 bands. Based on the MS images, an MLP is used for per pixel material classification whereas objects are detected based on RGB images. The results show reasonable material classification performance of our approach with a limited amount of training data and straightforward image registration and ML technologies.  The results also show that the NIR/SWIR spectrum is decisive for material classification in industrial/bulky waste and a subset of bands show a similar performance to the usage of all bands.

Experiments with state of the art object detection models have revealed that these methods show an unsatisfactory performance when applied to scenes of bulky waste heaps. Further challenges can arise from increased distance and interference from dust and dirt when applied to a setting in operational use.

Further experiments will be carried out to determine the decisive spectrums for material classification and narrow down the number of spectral bands.

An innovative, generic and cost-effective solution for autonomous control of hydraulic heavy machines for sorting of bulky waste is being developed. This provides an easier alternative to the traditional way of automating such machines, using cost-effective cameras and artificial intelligence-based controllers. The robot's state is estimated from camera images and the controller will learn from the data generated by moving the machines in manual control mode. This method eliminates the need for retrofitting joint encoders or explicitly modeling kinematics or dynamics of the robot.

This control concept is still in its initial stages of development and testing. Different methods are being tried and tested for each of the modules. The goal is to implement and test the concept for two different systems including an excavator and a crane.

Future work will investigate if detection and segmentation models which utilize multispectral data can make use of the additional information. These models may also directly incorporate material classification. In the chase of segmentation models this may also make a separate material classification model unnecessary. Additionally, a hydraulic crane in a recycling factory will be retrofitted with sensors and actuator and the described methods for autonomous control of hydraulic heavy machines will be deployed in a real setting. Finally, for the whole robotic system for autonomous coarse waste recycling a life cycle assessment will be carried out.

**ACKNOWLEDGEMENTS**

This work was supported by the Federal Republic of Germany, Federal Ministry for the Environment, Nature Conservation, Nuclear Safety and Consumer Protection based on a decision by the German Bundestag, grant no. 67KI21013A/B/C.

Special thanks to Mark Carson for his valuable proofreading of this article.